\documentclass{segabs}

\begin{document}

\title{Facies classification from well logs using an inception convolutional network}

\renewcommand{\thefootnote}{\fnsymbol{footnote}}

\author{Valentin Tschannen\footnotemark[2]\footnotemark[3], Matthias Delescluse\footnotemark[3], Mathieu Rodriguez\footnotemark[3] and Janis Keuper\footnotemark[2]\newline
\footnotemark[2]Fraunhofer ITWM, Kaiserslautern, Germany \newline
\footnotemark[3]\'{E}cole Normale Sup\'{e}rieure, Paris, France}

\footer{SEG}
\lefthead{Tschannen et al.}
\righthead{Automated facies classification from well logs}

\maketitle

\begin{abstract}
The idea to use automated algorithms to determine geological facies from well logs is not new (see e.g \cite{Busch_Fortney_Berry_1987}; \cite{rabaute1998inferring}) but the recent and dramatic increase in research in the field of machine learning makes it a good time to revisit the topic. Following an exercise proposed by \cite{Dubois:2007:CFA:1238149.1238451} and \cite{hall2016facies} we employ a modern type of deep convolutional network, called \textit{inception network} \cite[]{szegedy2015going}, to tackle the supervised classification task and we discuss the methodological limits of such problem as well as further research opportunities.
\end{abstract}

\section{Introduction}
Facies are used by geologists to group together body of rocks with similar characteristics in order to facilitate the study of a basin of interest. Their definition is rather subjective as it depends on the attributes we choose for the classification. One may for instance focus on biological differences by looking at the type of shells present in the samples or we may wish to emphasis petrological characteristics by accounting for the granulometry and the mineralogy. In the case of Oil$\&$Gas reservoirs, porosity and permeability are critical properties to determine since they give indications about the potential volume of fluids that might be stored in a rock and how they will flow during production. We can therefore expect that grains size, shape and density as well as the depositional and compaction history of the rocks will be a dominant factor for the categorization. While the main source of information for defining those facies comes from the observation of core samples under visible and x-ray light, we also have a variety of well log recordings at our disposal. By measuring the acoustic and electrical responses as well as the nuclear radiations of the drilled medium, we can infer properties about its rock matrix and fluid content and indirectly relate them to the porosity, permeability or fluid saturation of the rocks.

Classifying high dimensional data into groups is one of the main branch of the popular field of machine learning. Among the vast panel of methods, current attention is mainly received by so called \textit{deep neural networks}. Learning from experience, those algorithms are able to discover abstract representations and to understand the data in terms of a hierarchy of concepts. They have shown impressive results in a vast panel of supervised classification problems \cite[]{lecun2015deep}.

\plot*{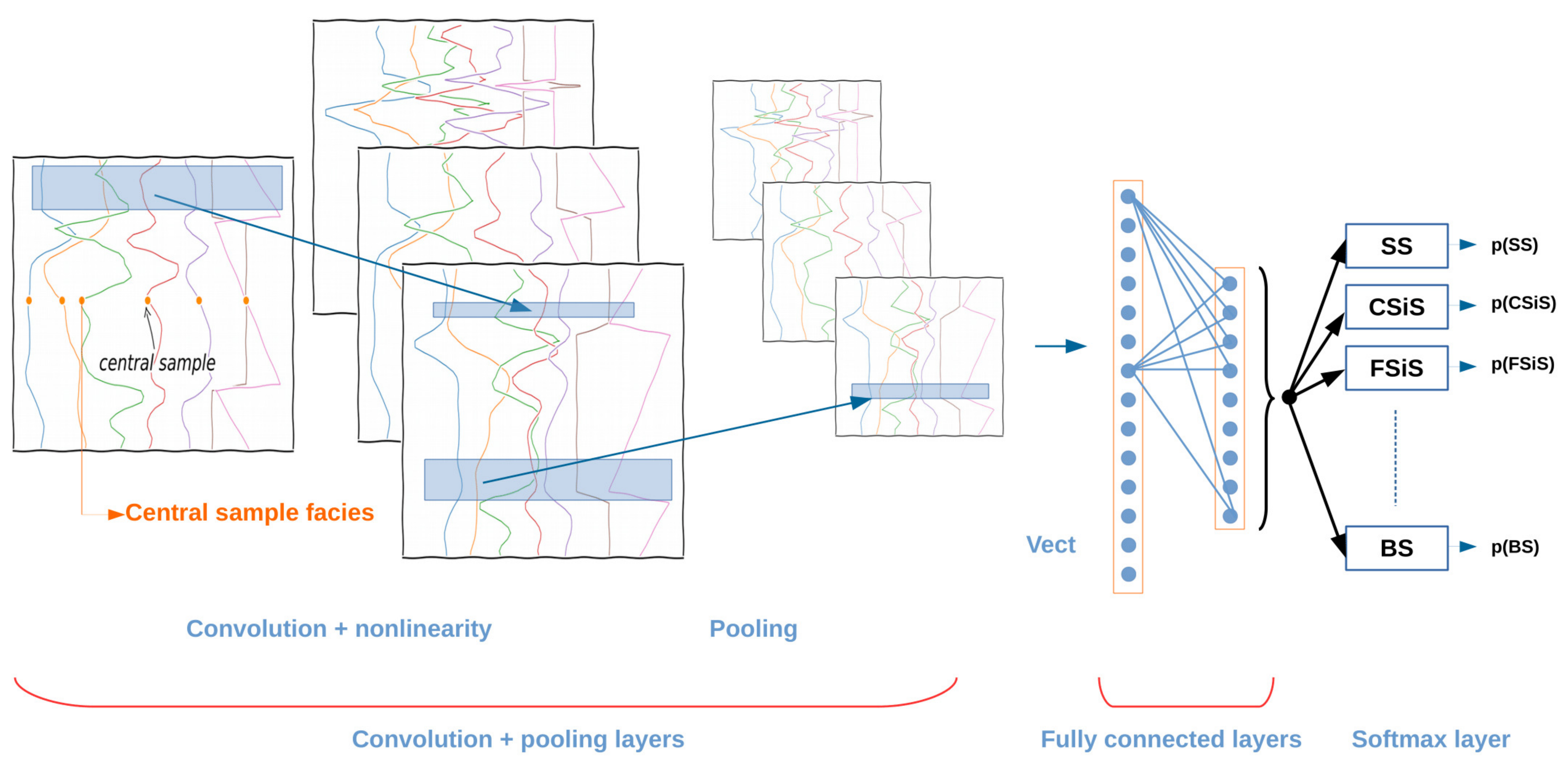}{width=0.8\textwidth}
{Schematic architecture of a 1 dimensional ConvNet. A short depth-window is extracted from the logs around an example of facies. Filters in the convolution layers are 2-dimensional, in measured-depth (meters or feet) and in channels (number of logs per well) but applied only along depth. The number of filters in the first layer will determine the number of input data to the second layer. Pooling operations reduce the length of the logs by dropping every other samples. On input to the fully connected layers, all filtered well logs are concatenated into a large 1-dimensional series. The softmax layer outputs a vector whose elements form a discrete probability distribution of the facies.}

\section*{method}
 
\subsection{Convolutional networks}
Deep networks have recently become the method of choice to solve problems in the fields of computer vision and speech recognition. Their general architecture can be seen as a sequence of layers connected to each other trough a non-linear differentiable function. Each layer is composed of a number of units that independently apply a simple affine transform on their input. The power of such approach resides in the way the coefficients of the transforms are set. Rather than being manually engineered they are given the freedom to adapt to the data by progressively learning from examples. Stacking several layers is a key in the success of those algorithms. Due to the non-linearity applied in between each layers, the learnt coefficients tend to be sensitive to progressively more abstract and complex features in the data as we go deeper in the network. The global perspective of those algorithms however has some inconveniences. They require a huge number of parameters and are difficult to train when working on real world datasets \cite[]{726791}.

To overcome those drawbacks, early researchers such as \cite{726791} proposed to restrict the so called receptive fields of each units to localized regions of the data. By making the observation that the world is compositional, they argued that instead of using the entire input at once to learn the coefficients, units should rather process local groups of samples in a sliding manner. This particular class of networks are named convolutional networks (ConvNets). In addition of being computationally and memory effective (each unit now posses only as many parameters as the size of its receptive field) it also appears to be a more robust way to proceed when it comes to field recordings. In the shallow part, by looking at localised regions of the data, the units typically learn wavelet-like convolutional filters that are useful in detecting basic features. The deeper units will take advantage of the simple feature detectors provided by the previous layers to make their own advanced detectors. Furthermore, by using zoom-out operations (referred in the literature as pooling) in-between the layers, deeper units will look at the data in a progressively more global way \cite[]{lecun2015deep}. As an example, if we assume the gradient of the well logs to be relevant in our problem, it is likely that knowing the gradient over the entire logs will be necessary. Rather than having to learn a large filter made of the concatenation of many differential operators, it is easier to simply learn one operator and apply it in a convolutional manner. 

ConvNets are typically designed in a reversed pyramid manner by increasing the number of units as we go to deeper layers. This will progressively transform the original data into a very high dimensional, non-linear space, where clustering, classifications or regressions can be effectively conducted.   

\subsection{Classification}
In this work we are interested in employing a ConvNet to solve a supervised classification problem. Using training data coming from wells where the facies sequences were already determined by geologists, we train the network to predict facies series that accurately match the experts' interpretation. To this end, in addition to the convolutional and pooling layers described in the previous section, we also need to append fully connected (fc) layers and an output layer. Unlike the convolutional layers, fc layers have access to the entire input at once. Since they come after the deepest convolutional layer, they will be fed with already abstracted individual feature detectors. With the help of additional non-linearities between them, their role is to appropriately combine the abstracted features together in order to solve the supervised task. The output layer is converting the information fed by the fc layers (usually) as a discrete probability distribution of the facies. At each iteration, a random subset of the examples are sent to the network, and using an appropriate objective function, such as the cross-entropy, we compare the predicted labels to the training labels. A global error term is computed and back-propagated though the network to update all the coefficients, in order to minimize the classification error. This approach is called the stochastic gradient descent \cite[]{726791}. If the network was properly trained, it should now be able to generalise to new wells where we do not know in advance the facies sequence.

\subsection{The inception modules}
\cite{szegedy2015going} proposed an interesting modification to the convolutional layers. Instead of having a homogeneous setting, they split the layers into four distinct paths (Figure~\ref{fig:inception-1-eps-converted-to.pdf}) and they introduce the 1$\times$1 convolutions as a cost-effective and efficient feature augmentation technique. The 1$\times$1 convolutions can be seen as a weighting of the different input filters, producing many possible linear combinations that the network can choose from in later layers, while keeping the number of outputs reasonably small. They are also followed by a non-linear activation which further improves the generalization power of the network. 

The layout of the inception module can be seen in Figure~\ref{fig:inception-1-eps-converted-to.pdf}. The data follows each of the four path in parallel before being concatenated at the output. Since it is unclear whether the low frequency or the hight frequency (or both) information contained in the well logs will dominate the learning process, we better let the algorithm decides on its own. As we stack those modules on top of each other, the network progressively learns more and more abstract features from the data which can then be fed to the softmax classifier.

\plot{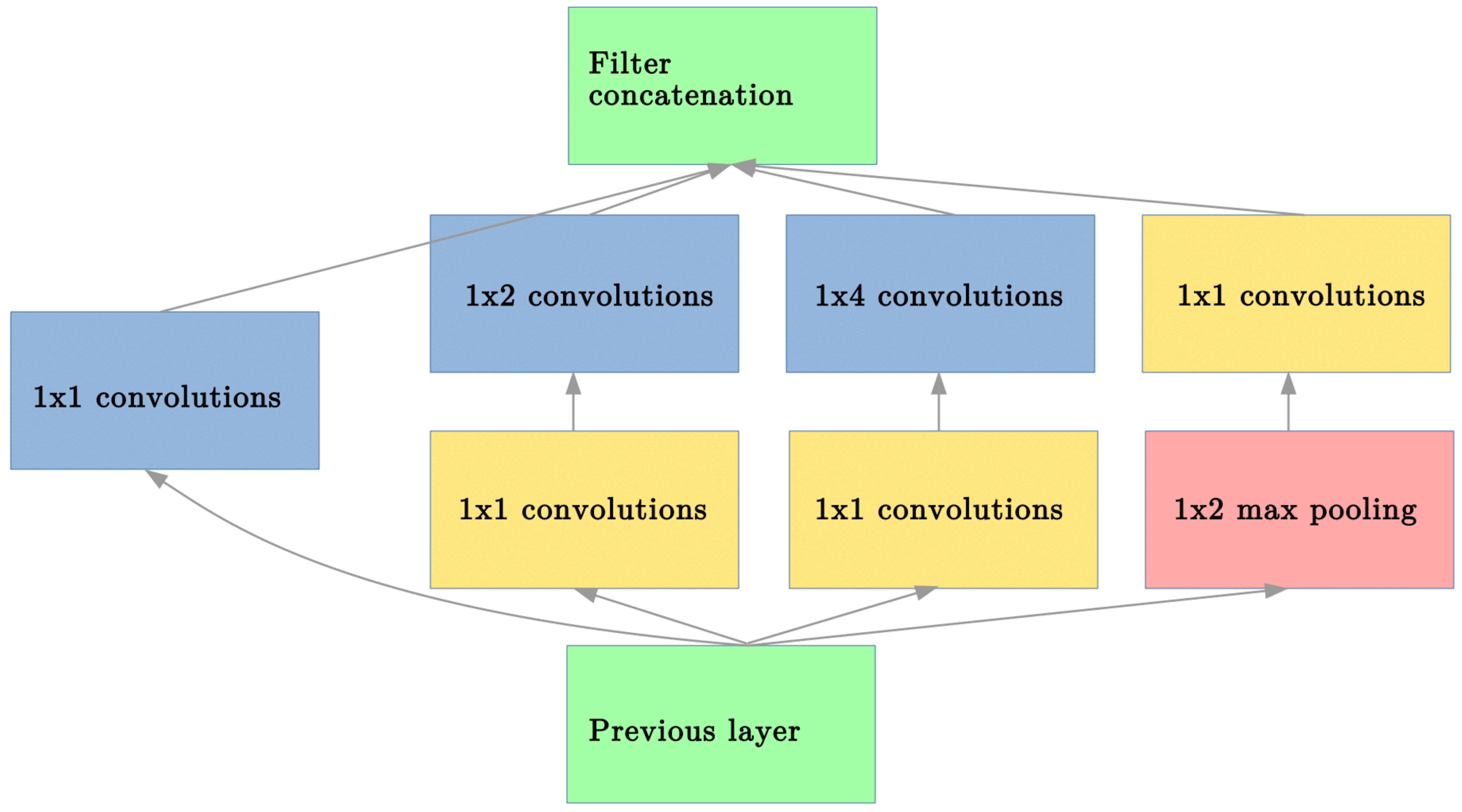}{width=\columnwidth}
{Inception module architecture after \cite{szegedy2015going}. On the left, the 1$\times$1 convolution will preserve the vertical resolution of the log sequences. The small kernel convolution will be more sensitive to high frequency informations. The large kernel convolution will be more sensitive to low frequency information. Finally, on the right side, a pooling followed by a 1x1 convolution will perform a sort of low-frequency filtering of the logs, in order to progressively look at more spatially averaged features.}

\subsection{Regularization techniques}
The biggest challenge in teaching a deep network lies in preventing over-fitting, who’s symptom is a large drop between the training and the testing performances. Rather than learning parameters that capture the nature of the data, it is often observed that a network simply memorizes the training examples. Hence, this results in a very high prediction accuracy on the examples it has already seen, but in poor generalization performances. The most important is to make sure that the problem we want to solve is well-posed and that we have enough training samples. Additionally, many preconditioning and regularization techniques have been developed. Common practice involve a standardization of the input data, an appropriate choice of objective function with the possible addition of penalty terms and the use of dropout \cite[]{srivastava2014dropout} to enforce redundancy in the learnt filters weights. Hyper-parameters, such as the learning rate (step size of the gradient descent) should be carefully tuned by using subsets of the training examples for blind validation. Due to the random initialization of the network and the stochastic nature of the learning phase, results will vary from run to run, and one should aim to design a stable algorithm.

\section{Experiment}
The data we used and the experiment settings originate from a class taught at the University of Kansas \cite[]{bohling2003integrated}; \cite[]{Dubois:2007:CFA:1238149.1238451}. An up-to-date implementation of our method with the library \texttt{Tensorflow} can be found on Github\footnote{https://github.com/vts21/2016-ml-contest/tree/master/itwm}. A total number of 11 wells were supplied, each containing 7 logs and a corresponding rock facies series. Out of those 11 wells, 9 were used for training the network and the remaining 2 for a blind evaluation. Among the logs, 5 come from wireline measurements sampled every 1.5 meters (gamma ray, resistivity logging, photoelectric effect, neutron-density porosity difference and average neutron-density porosity) plus an additional 2 geological constraints (non-marine versus marine indicator and relative position) derived from knowledge of the reservoir area. Moreover, we remark that the 2 neutron-density logs are computed using mineralogy dependent coefficients, which may be an additional source of uncertainty.

From observations of core samples, geologists determined that the stratigraphy could be described by 9 different facies: Nonmarine sandstone (SS), Nonmarine coarse siltstone (CSiS), Nonmarine fine siltstone (FSiS), Marine siltstone and shale (SiSh), Mudstone (limestone - MS), Wackestone (limestone - WS), Dolomite (D), Packstone-grainstone (limestone - PS) and Phylloid-algal bafflestone (limestone - BS).

We show in Figure~\ref{fig:logplot_574_bigger-eps-converted-to.pdf} the results obtained for one of the two blind wells and we assess the quality of the machine's prediction with respect to the geologists' classification in term of the F1 score (Figure~\ref{fig:cm_536-eps-converted-to.pdf}). Visually, the predicted stratigraphy seems reasonable in the first order, as it looks like a median filtered version of the man-made solution. However, an average F1 score of 0.574 and the many observable short intervals in the lithofacies sequence presenting dissimilarities, bring out the difficulties of the network to match the geologists' answer with a high resolution. For instance, if we focus on the packestone layer (PS, in purple) predicted by the machine with a fairly high confidence between 2990m and 3005m, we see a drop in the probabilities at the borders indicating uncertainties in the exact location of the transitions from the above siltstone and the coming wackestone. We also observe that the very fine layers (between 1.5m and 3m thick) of dolomite, mudestone and wackstone found inside the main packestone layer by the geologists were simply not recognized by the machine.

\plot{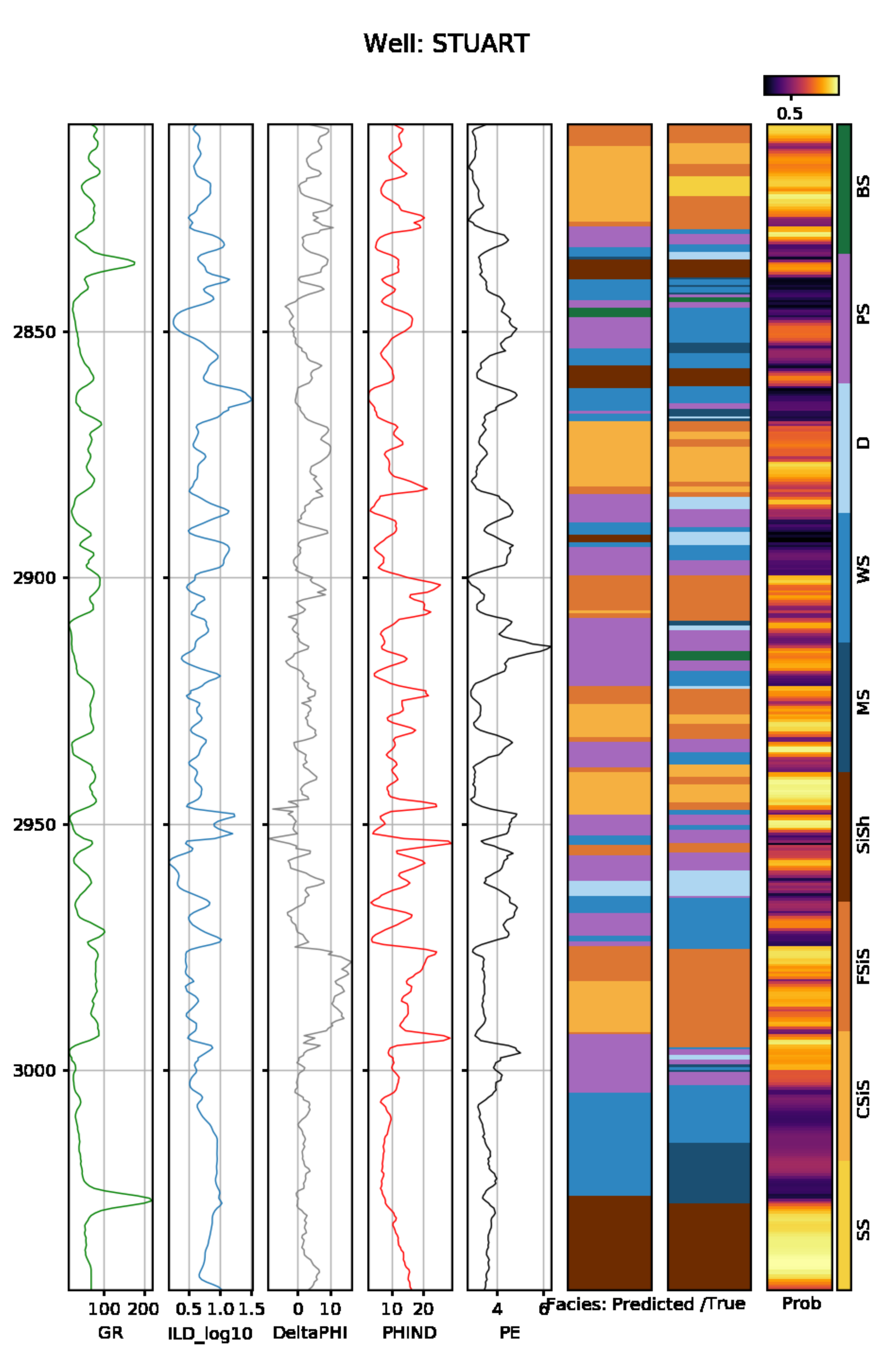}{width=.9\columnwidth}
{Results of the classification on a blind well. The five first plots show the measured logs as a function of depth (in meters). The two facies series represent the prediction by the machine on the left and the solution proposed by the geologists on the right. The last plot gives the probability with which the machine selected the facies. Bright yellow colors indicate a high certitude (above 70$\%$) while the dark colors indicate uncertain regions (below 50$\%$). The colorbar on the very right gives the correspondences to the facies. (Plot modified from \cite{doi:10.1190/tle34040440.1}).}

\plot{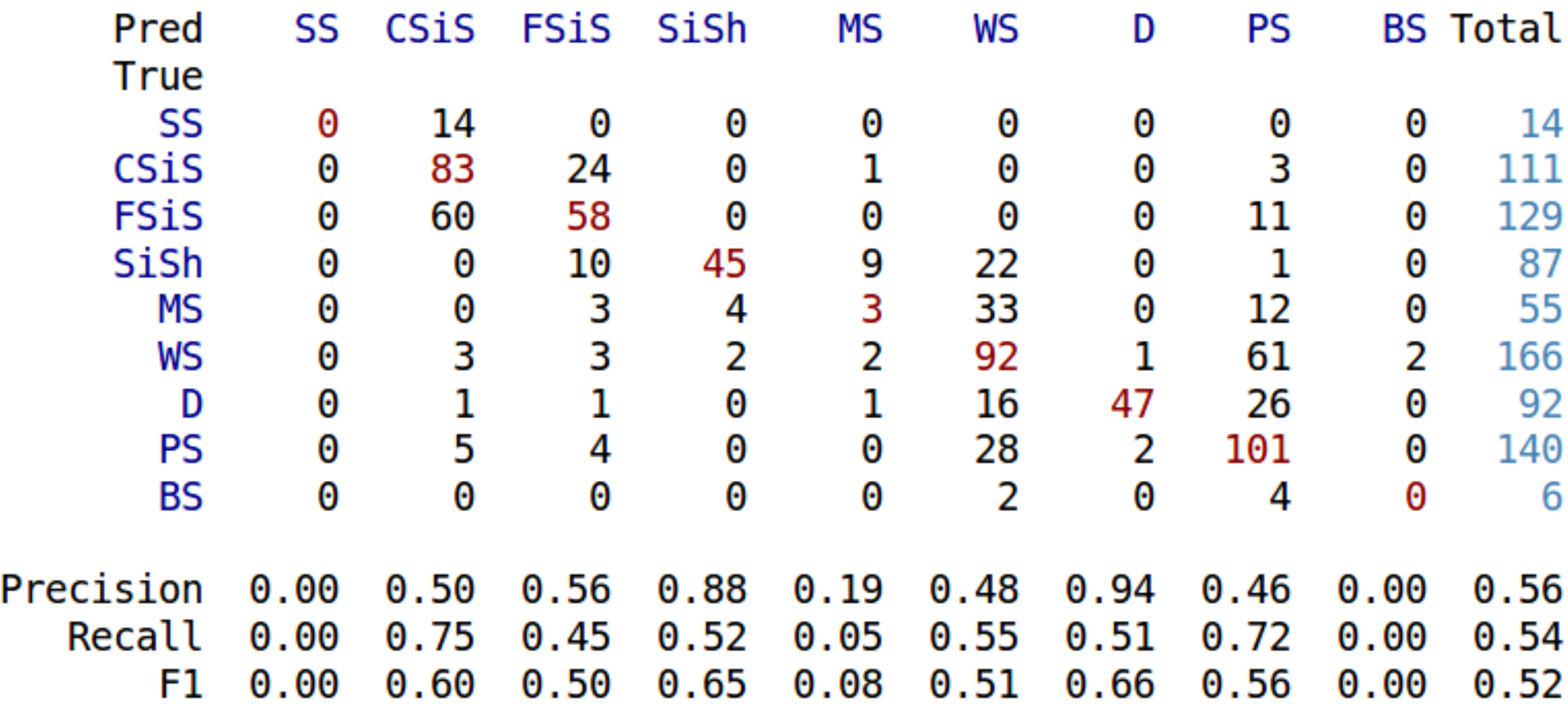}{width=\columnwidth}
{Confusion matrix for the results shown in Figure \ref{fig:logplot_574_bigger-eps-converted-to.pdf}. The last column (in blue) shows the total number of facies present in the geologists' interpretation. The diagonal (in red) gives the number of facies that were correctly classified by the machine. Off diagonal terms represent the miss-classifications of the machine. E.g, the number in the first row, second column indicates that the machine classified 14 samples as coarse siltstone (CSiS) whereas the geologist classified them as sandstone (SS). Precision is the ratio of true positives to all predicted positives. Recall is the ratio of true positives to all actual positives. The F1 metric weights recall and precision equally.}

\section{Discussion}
The results published by \cite{hall2016facies} and the SEG show that all the \textit{deep learning} methods gave F1 scores below 0.60 and that the \textit{decision tree} like methods fell below 0.65. For a machine learning contest those scores are surprisingly bad, and it is also interesting to note that deep learning approaches do not dominate. Concerning the later observation, we believe that the poorer performances of deep learning is due to a lack of data, as the power of those algorithms comes at the cost of supplying a profusion of training examples. As for the poor match in terms of F1 score, \cite{Dubois:2007:CFA:1238149.1238451} point out that the facies zones are more of a continuum rather than a clearly discrete sequence. This means that neighbour classes can be very similar, and a clear frontier does not exist. Moreover, human interpretation is non-unique and subject to errors, which is an additional challenge for geoscientific applications of machine learning as the ground truth is never know for certain. 

As previously mentioned, geologists used visual observations of core samples as well as general knowledge about the area to determine their sequence. On the other hand the network was only given access to well data. Since the resolution at the log scale is lower than the resolution at the core scale, this alone should explain why the very thin layers labelled by the geologists are not always recovered by the algorithm. Besides, it is not obvious that the distinctive visual criteria chosen by the experts will also appear in our 7 logs. This could mean that the boundaries drawn by the geologists cannot be completely recovered by the machine, making the problem we want to solve ill-posed. As an example, the confusion matrix in Figure~\ref{fig:cm_536-eps-converted-to.pdf} highlights the difficulty encountered by the network to separate the coarse siltstone (CSiS) and the fine siltstone (FSiS). To distinguish between the two clastic sedimentary rocks, petrologist are measuring the grain size. Given the well logs available, it is unlikely that the machine can match the power resolution of the human eye even considering the differences in porosity. It would have been informative to have a sonic log in addition, as grain size, and so the softness of the rock, strongly influences the acoustic propagation. 

An other factor to take into account is the proportion occupied by the different facies in the training and validation data. Figure~\ref{fig:barplot_facies_count-eps-converted-to.pdf} reveals that the algorithm saw almost seven times more examples of coarse sisltstones (CSiS) than dolomites (D). However, despite this, the confusion matrix indicates that classification performances for the dolomites reached an F1 score of 0.66 against 0.60 for the coarse siltstones. Therefore, it seems that the main problem is not the quantity of data but rather the resolution limitations discussed in the previous paragraph. 

\plot{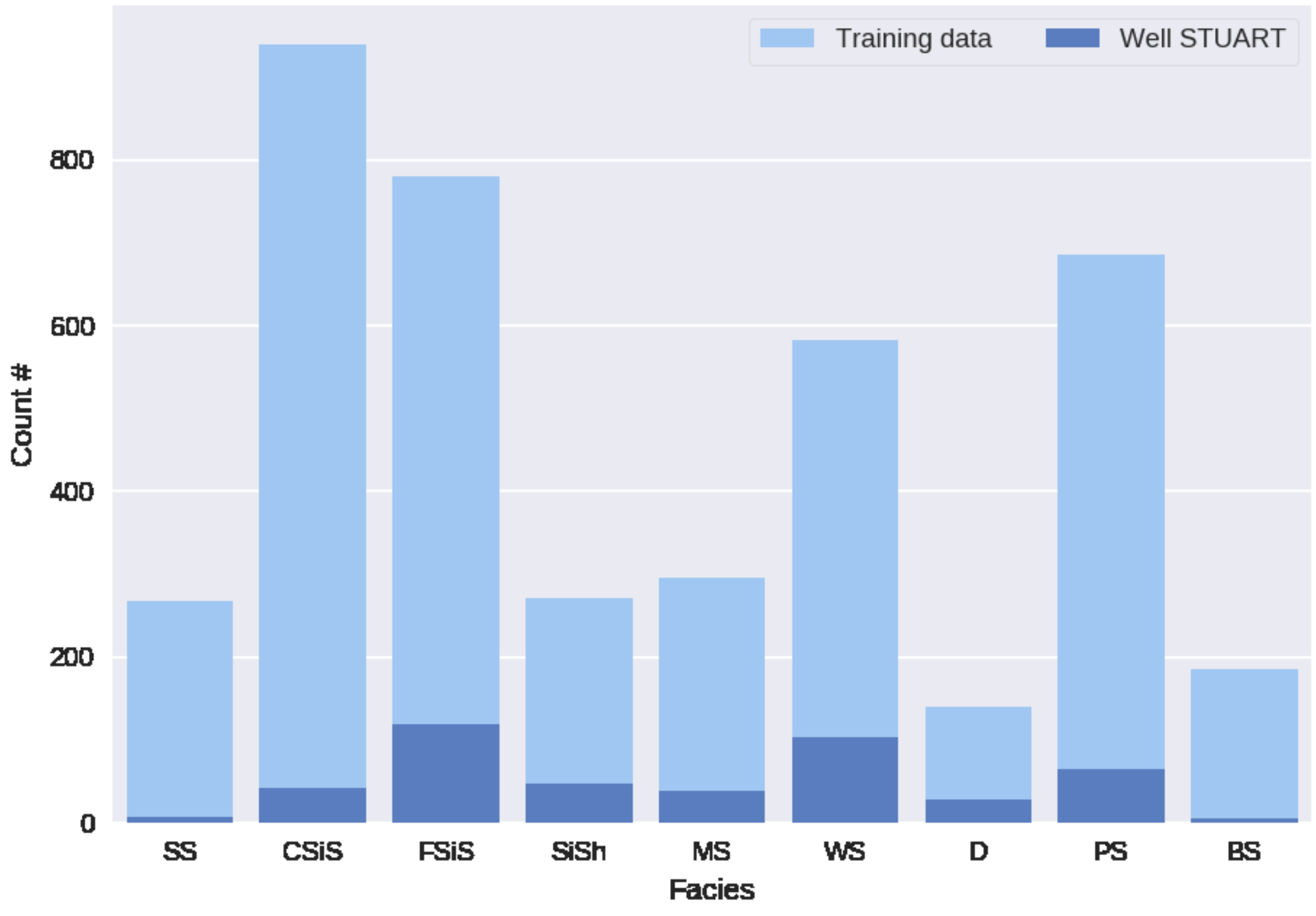}{width=\columnwidth}
{Bar-plot representing the total number of occurrences of each of the 9 facies for both the training data and the geologists interpretation for one of the blind wells (Stuart).}

As noted again by \cite{Dubois:2007:CFA:1238149.1238451}, being able to correctly classify within one neighbour facies is just as good as being correct, and since sources of information given to the geologists and the machine were different, we believe that evaluating the success of the exercise in terms of F1 score alone is of little interest. The machine's classification is not in competition with the humans' one but should rather be evaluated as a complementary information. Discussions with geologists who know the reservoir will indicate whether the global machine interpretation is nevertheless interesting.

\section{Conclusion}
We proposed to train an inception network to determine the stratigraphy of a reservoir from well measurements. By extracting short sequences from the logs around examples of facies, the algorithm progressively learns a facies - dependent parametrization of the data. We evaluated the prediction performances on blind wells and provided visual and statistical information about the results. The machine's answer is satisfying in the first order but, being deprived of higher resolution core samples data and of additional well measurements, it failed at reproducing the work of the geologists with a high accuracy. In future work we will investigate a form of unsupervised classification called clustering. Instead of teaching our network to recognize the facies chosen by the geoscientists, we may give it the freedom to come up with its own classes and analyse the similarities and differences with the human interpretation. Additionally, it shall be interesting to work with hybrid networks which accept different types of data, from the core scale to the seismic scale (see e.g \cite{Wang2012MethodologyOO}).




\bibliographystyle{seg}  
\bibliography{facies_class_2016_arxiv}

\end{document}